\documentclass{article}

\usepackage[preprint]{neurips_2020}

\usepackage[utf8]{inputenc} % allow utf-8 input
\usepackage[T1]{fontenc}    % use 8-bit T1 fonts
\usepackage{hyperref}       % hyperlinks
\usepackage{url}            % simple URL typesetting
\usepackage{booktabs}       % professional-quality tables
\usepackage{amsfonts}       % blackboard math symbols
\usepackage{nicefrac}       % compact symbols for 1/2, etc.
\usepackage{microtype}      % microtypography
\usepackage[linecolor=lightgray,bordercolor=lightgray,backgroundcolor=white,textsize=footnotesize]{todonotes}
\usepackage{caption}
\usepackage{subcaption}
\usepackage[capitalise, noabbrev]{cleveref} % \cref{fig:label}
\usepackage{pst-node}

\title{Mathematical Reasoning via Self-supervised Skip-tree Training}

\author{%
  Markus N.~Rabe \\
  Google Research \\
  \texttt{mrabe@google.com}
  \And
  Dennis Lee \\
  Google Research \\
  \texttt{ldennis@google.com}
  \And
  Kshitij Bansal \\
  Google Research \\
  \texttt{kbk@google.com}
  \And
  Christian Szegedy \\
  Google Research \\
  \texttt{szegedy@google.com}
}

\begin{document}

\maketitle

\begin{abstract}
We examine whether self-supervised language modeling applied to mathematical formulas enables logical reasoning.
We suggest several logical reasoning tasks that can be used to evaluate language models trained on formal mathematical statements, such as type inference, suggesting missing assumptions and completing equalities.
To train language models for formal mathematics, we propose a novel skip-tree task.
We find that models trained on the skip-tree task show surprisingly strong mathematical reasoning abilities, and outperform models trained on standard skip-sequence tasks.
We also analyze the models' ability to formulate new conjectures by measuring how often the predictions are provable and useful in other proofs.
%that do not fit the ground truth or any training data turn out to be true and useful statements.
\end{abstract}

%%%%%%%%%%%%%%%%%%%%%%%%%%%%%%%%%%%%
\section{Introduction}
%%%%%%%%%%%%%%%%%%%%%%%%%%%%%%%%%%%%

Language modeling using Transformers~\citep{vaswani2017attention} has been hugely successful for applications like translation and text generation.
Models like GPT-2 are able to generate impressive news articles and stories given just an abstract~\citep{radford2019gpt}.
These models are usually first trained on a proxy task, such as predicting missing words in the case of BERT~\citep{devlin2018bert}, before fine tuning the models on more specific (downstream) tasks such as machine translation and question-answering.
The proxy tasks are not reliant on labeled data, and thus can be trained on large corpora of unlabeled data.
Even the models trained on the proxy tasks alone, have shown impressive language understanding~\citep{brown2020language}.

Prior work in deep learning for mathematics has focused on learning directly on logical reasoning tasks, such as predicting the proof steps or premises or assignments.
These approaches require \emph{labeled} data, which is hard to come by and typically very limited in size.
In this work, we apply the paradigms of language modeling to formal mathematics and define proxy tasks on \emph{unlabeled} mathematical expressions that allows us to use much more data.
% Instead of training directly for a logical reasoning task like most previous works in deep learning for mathematics, we simply start from the HOList dataset and define proxy tasks similar to language modeling.
We start with the HOList dataset~\citep{bansal2019holist}, which spans a wide range of mathematical topics, including topology, multivariate calculus, real and complex analysis, geometric algebra, and measure theory, formalized in the HOL Light proof assistant~\citep{Harrison96}.
We find that training a language model on all mathematical expressions in this dataset leads to surprisingly strong mathematical reasoning capabilities.

For training language models on formal mathematics, we propose a novel skip-tree task. 
The skip-tree tasks is a specialization of the skip-sequence task that respects the tree structure of expressions.
% The only difference is that the skip-tree task respects the tree structure of expressions, 
We show that models trained on the skip-tree task significantly outperform those trained on the standard skip-sequence task.

%We then study the logical reasoning abilities of our language models.
Reasoning can refer to a wide range of abilities, and thus we measure the mathematical reasoning abilities of language models on a variety of tasks, including mechanical derivations, such as type inference, and also creative tasks, such as predicting under which assumptions a statement is true.
In contrast to most works in natural language modeling, we do not fine-tune the models to the evaluation (downstream) tasks, as we want to study what reasoning capabilities can be acquired just through language modeling proxy tasks.
% what the language models learned only through the proxy task.
% One of the big challenges in applying deep learning to mathematics is that the data sets are relatively few and small.
% Similar to natural language modeling, we define a proxy task to learn that does not required labeled data, and 

% There are few differences to natural language modeling that are worth highlighting.
% First, for formal mathematics, the data sets are relatively few and small.
% Applying deep learning to mathematics is a relatively new field, and there is a shortage of labeled training data.
% To address that, we extract a large amount of training data automatically from a set of mathematical statements that were created and used by humans in a proof assistant.
% Second, we do not fine-tune our models, and instead evaluate on the reasoning tasks directly.

An advantage of formal language compared to natural language is that we can attempt to automatically evaluate statements.
That is, even if the language models fail to predict the ground truth, the statements they predicted might still be true and useful.
We evaluate these \emph{conjectures} by attempting to prove them and checking if they are can be used in the context of other proofs.
% For instance, one can attempt to verify if the generated statements are true, or if they are useful to prove other statements.
% We run these evaluations, in addition to looking for exact matches, and qualitative manual evaluation.

Our contributions are as follows:
\begin{enumerate}
    \item We introduce several evaluation tasks that test logical reasoning abilities.
    \item We introduce a new skip-tree language modeling task that outperforms skip-sequence approaches in our evaluation on the logical reasoning tasks.
    \item We show that language modeling on mathematical formulas results in surprisingly strong logical reasoning capabilities.
    \item We suggest a way to create and evaluate mathematical conjectures with language models.
\end{enumerate}

The remainder of this paper is structured as follows:
First, we review related work on language modeling and deep learning for mathematics in Section~\ref{sec:related}.
Then, in Section~\ref{sec:dataset} we discuss the source corpus of formal mathematical statements from which we generate our training data.
In Section~\ref{sec:skiptree}, we present our novel language modeling task for formal languages, as well as several variations that we used in our ablation studies.
We present the evaluation tasks in Section~\ref{sec:evaluationtasks}, present our experimental findings in Section~\ref{sec:results}, and conclude in Section~\ref{sec:conclusion}.

%%%%%%%%%%%%%%%%%%%%%%%%%%%%%%%%%%%%%%%%%%%%%%%%%%%%%%%%%%%%%%%%%%%%%%%%
\section{Related work}
\label{sec:related}
%%%%%%%%%%%%%%%%%%%%%%%%%%%%%%%%%%%%%%%%%%%%%%%%%%%%%%%%%%%%%%%%%%%%%%%%

Recently, we have seen a series of rapid improvements in language modeling stemming from better pretraining tasks~\citep{devlin2018bert,zhang2019pegasus,song2019mass,dong2019unified,raffel2019exploring,conneau2019cross}.
BERT \citep{devlin2018bert} is a pretraining task for Transformers~\citep{vaswani2017attention}, which masks out a certain fraction of the input tokens that the model then has to predict.
UniLM uses multiple pretraining tasks~\citep{dong2019unified}.
One of them is a sequence-to-sequence task; to predict the next sentence from the previous sentence.
MASS and SpanBERT consider a generalized sequence-to-sequence pretraining task, which is to predict a masked out subsequence of the input~\citep{song2019mass,joshi2020spanbert}.
% SpanBERT additionally considers a span boundary objective, which is to predict the masked out subsequence only from the tokens adjacent to the missing subsequence~\citep{joshi2020spanbert}.
However, both MASS and SpanBERT reveal the length of the sequence to predict as they replace it by a number of mask tokens equal to the length of the sequence.

T5 introduced a generalization of sequence-to-sequence pretraining tasks that is crucial to our work~\citep{raffel2019exploring}.
They replace the subsequence (or multiple subsequences) to be predicted by a single token (not a number of mask tokens equal to the length of the subsequence, as in MASS).
% Further, T5 allows multiple subsequences to be masked out and predicted.
\citep{zhang2019pegasus} additionally exploit the sentence structure of natural language.
They suggest the pretraining task Pegasus, which masks out entire sentences of a given text, and additionally masks out randomly selected tokens in the remaining text (or alternatively replace them by other tokens).
In a similar way Pegasus' exploitation of the sentence structure of natural language, our skip-tree task exploits the tree structure of formal expressions.
\citep{zhang2019pegasus} also suggest sampling the sentences to be masked with the help of ROUGE1-F1~\citep{lin2004rouge}.

% Note that in contrast to most works for natural language, we do not fine tune our models on the evaluation (downstream) tasks.
% Instead we intentionally evaluate them on out-of-distribution evaluation tasks to analyze what they have learned.

We work with the HOList dataset by~\cite{bansal2019holist}, which is closely related to the Flyspeck dataset by~\cite{kaliszyk2014learning}.
There are other datasets which might be suitable for our approach as well, including proofs extracted from HOL4~\citep{gauthier2017tactictoe}, and from Coq~\citep{huang2018gamepad,yang2019learning,proverbot2019}.  % proverbot2019

Most previous works that apply sequence-to-sequence models to logics have focused on specific logical tasks in \emph{supervised} training settings.
In contrast, we train language models on an unsupervised proxy task that does not require labeled data and can thus be applied to almost any source of mathematical expressions. % and evaluate them on several logical reasoning tasks that are substantially different from the training task.
\cite{lample2019symbolicmathematics} use a Transformer model for symbolic integration.
They train their model directly on the reasoning task they want to learn, and their approach requires that the inverse of the prediction task can be computed effectively with classical algorithms.
% Also, our dataset spans a much wider range of mathematical theories. %, a much larger vocabulary and variety of inputs.
% We imagine that the language modeling approach explored in this paper, could be used as a pretraining task for symbolic integration.
% \citep{lee2020mathematical} show that graph neural networks can predict embeddings of statements that result from applying rewrite operations to a statement, even several steps of rewrite operations, and that this task can be learned in purely self-supervised manner.
\cite{finkbeiner2020teaching} explore the generalization properties of the Transformer architecture predicting the solutions to SAT formulas and temporal logic, but require a data generator that can solve formulas, which is currently not feasible for higher-order logic.
\cite{piotrowski2019can} train RNNs on individual logical reasoning steps, such as substitutions, using a dataset of rewrites on polynomials extracted from Prover9.
\cite{wang2018first} translate between synthetic descriptions in natural language and formal mathematics on a dataset generated with Mizar.

Self-supervised training techniques for formal mathematics have received much less attention.
\citet{wang2020exploration} apply recent unsupervised translation techniques by \citet{lample2018unsupervised} to align formal and informal techniques.
They that they perform considerably worse than supervised techniques.
Very recently, \citet{urban2020neural} presented initial experiments on applying self-supervised language modeling to formal mathematics in order to produce conjectures.
Earlier statistical approaches to produce conjectures include~\citet{gauthier2016initial}.
Also, very recently, \citet{li2020modelling} applied language modeling to proofs of formal mathematics.

Transformer models for program understanding have focused on providing inductive biases in the architecture~\citep{shiv2019novel,Hellendoorn2020GREAT}, whereas this work suggests to use a modified language modeling proxy task.

Applying natural language techniques to formal mathematics has a long history.
Already in 2004, \cite{cairns2004informalising} applied information retrieval based on latent semantics to improve over search for keywords, and \citet{urban2004mptp} formulated the intention to learn from large amounts of data in formalized mathematics.

\section{Dataset}
\label{sec:dataset}
%%%%%%%%%%%%%%%%%%%%%%%%%%%%%%%%%%%%%%%%%%%%%%%%%%%%%%%%%%%%%%%%%%%%%%%%

We start from the HOList dataset introduced by \cite{bansal2019holist}.
The complete dataset includes 29465 theorems and their proofs.
We here consider only the ``core'' and ``complex'' datasets which comprise 18943 theorems, 637 definitions and 603,950 proof steps. These proof steps were extracted from the human proof logs.
The theorems and proofs were written (by humans) using the HOL Light proof assistant, and span various areas of mathematics such as set theory, arithmetic, linear algebra, topology, and multivariate complex analysis. % which checks the proofs for correctness.
% TODO: define goal/initial goal
The proofs contain a lot of intermediate goals which are the result of applying ``tactics'' on previous proof goals.
For example, one of the tactics is to rewrite the current proof goal with a set of equations selected from the theorem database. % or arithmetic simplification of formulas.

From this dataset we extract all theorem statements as well as all intermediate proof goals.
We use S-expressions to represent all statements.
For example, \texttt{(v bool x)} represents a boolean variable named \texttt{x}, and \texttt{(a (v (fun (bool) (bool)) f) (v bool x)} represents the function application $f(x)$ where $f$ is a function from bool to bool.
The S-expression syntax is thus very verbose, which can cause some expressions to not fit into the size constraints of our Transformer model.

\begin{figure}
    \centering
    \includegraphics[width=0.7\textwidth,trim=20 120 65 20,clip]{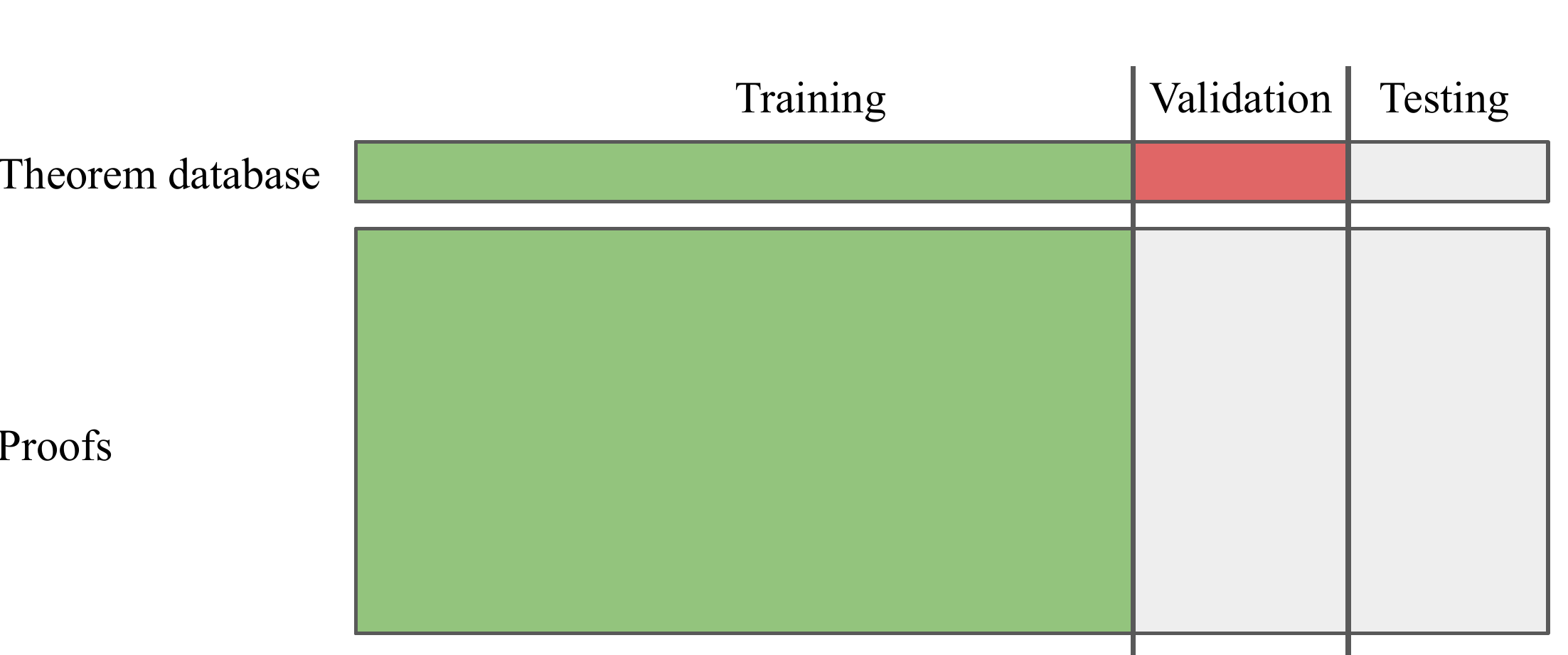}
    \caption{We use the theorems and proofs of the training split, marked in green, for training.
    For our evaluation tasks, we only use the theorems of the validation set, marked in red, to ensure that the model has never seen the statements from which the evaluation tasks are derived.}
    \label{fig:split}
\end{figure}

We use the same split into training/validation/testing data as defined in HOList.
The split is defined on the theorems, and the entire proof of each theorem is assigned to the same split as the theorem.
This means that we have used the proof of 11,655 theorems in the training split of the core and complex libraries.
This avoids partially revealing the proofs of theorems in the validation and test sets during training.
We derive all training data from the theorems and proofs in the training set, and use only the theorems (not the proofs) for the evaluation tasks.
This addresses the possibility that some proof steps for training theorems and for validation theorems might be shared.
In Figure~\ref{fig:split} we depict our choice of training and evaluation data.

%%%%%%%%%%%%%%%%%%%%%%%%%%%%%%%%%%%%%%%%%%%%%%%%%%%%%%%%%%%%%%%%%%%%%%%%
\section{Skip-tree Training}
\label{sec:skiptree}
%%%%%%%%%%%%%%%%%%%%%%%%%%%%%%%%%%%%%%%%%%%%%%%%%%%%%%%%%%%%%%%%%%%%%%%%

\begin{figure}
    \centering
    \includegraphics[width=\textwidth,trim=20 120 70 70,clip]{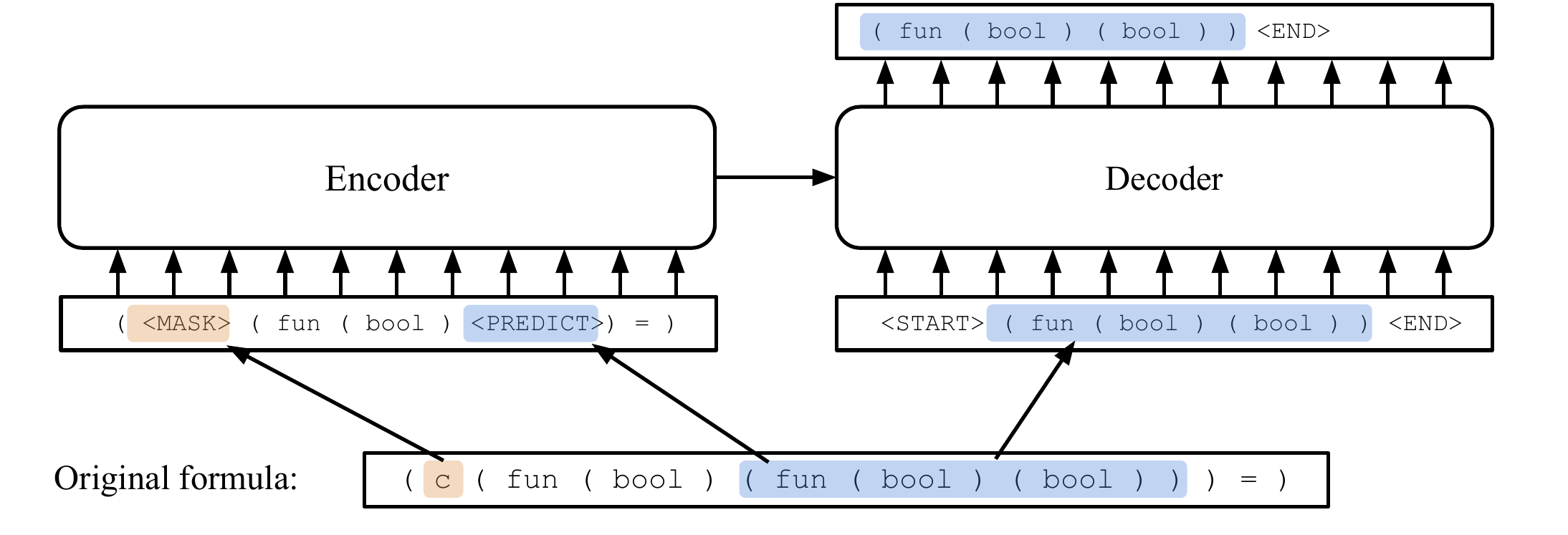}
    \caption{The skip-tree training task for the example of the equality operator on boolean constants (original formula).
    In this example we assume that a part of the type was sampled to be the subexpression to be predicted, and that subexpression \texttt{c} was sampled to be masked out additionally.
    Note the input to the decoder is shifted to the right, such that the next token prediction task yields the target sequence.}
    \label{fig:skiptree}
\end{figure}

In this section we define the skip-tree training task.
We
%assume that we can 
parse a given mathematical statement into a tree of subexpressions, and replace one of the subexpressions by a <PREDICT> token.
The task is to predict the subexpression replaced by <PREDICT>.
See Figure~\ref{fig:skiptree} for an example.

For training, the trees are converted back to a sequence of tokens; the target sequence is extended by a <START> token in the front and an <END> token in the back.
We exclude training examples where the output sequence is longer than the length of the decoder (512 tokens), and we cut off input sequences when they exceed the length of the encoder (1024 tokens).

\paragraph{Additional masked subexpressions.}
In addition to the subexpression to be masked out by \mbox{<PREDICT>}, we select $k=2$ subexpressions to be masked out by a different mask token <MASK>.
In contrast to the <PREDICT> token, we replace \emph{all occurrences} of these subexpressions by the <MASK> token.
Note that it can happen that the subexpressions we want to replace by the <MASK> tokens overlap with each other or with the subexpression replaced by the <PREDICT> token.
In this case, we give the highest preference to the <PREDICT> token, and then in decreasing order of size for the expression to be replaced by the <MASK> tokens.

The subexpressions masked by <MASK> do not have to be predicted.
They are only hidden to make the task harder and to make the model tolerant to having partial information.
A beneficial side effect of replacing some expressions by a <MASK> token is that the input sequences get substantially shorter and more mathematical expressions fit in the size constraints of the Transformer architecture.

\paragraph{Distributions of subexpressions.}
Sampling subexpressions uniformly at random results in very short sequences to be predicted: since our trees are mostly ternary, two thirds of the subexpressions are leaves.
Besides picking subexpressions uniformly at random, we thus experiment with weighting the subexpressions by the number of tokens they contain.
We refer to these variants as ``uniform'' and ``weighted''.
This results in a much more diverse set of expressions to be sampled.

\paragraph{Multiple samples per statement.} Since we started with a data source that is small compared datasets in natural language modeling, we use each mathematical statement from the training set to generate $n=100$ training examples.
Our initial data consists of about 360K intermediate statements from the proofs of 10K statements in the training split of the core and complex library of the HOList corpus.
To avoid duplicates, we sample the subexpressions that are replaced by a <PREDICT> token for each original formula \emph{without replacement}.

\subsection{Ablations}

To verify the design choices of the skip-tree training task we generated multiple variants of the training task and trained a model on each of them.

\vspace{-1mm}
\paragraph{No mask tokens.} To answer the question of whether it helps to mask out subexpressions besides the one to predict, we generated a dataset with $k=0$, called ``skip-tree (no <MASK>)''.

\vspace{-1mm}
\paragraph{Fewer samples per statement.} Instead of sampling many training examples from each formula, we could train on a fewer training examples for more epochs.
We generated a smaller version with $n=20$ of the skip-tree training data, which we call ``skip-tree (small)''.

\vspace{-1mm}
\paragraph{Skip-sequence.} MASS~\citep{song2019mass}, SpanBERT~\citep{joshi2020spanbert}, and T5~\citep{raffel2019exploring} pretrain their sequence-to-sequence natural language models by predicting subsequences of the tokens.
The skip-tree task is similar, but exploits our ability to parse the formulas as trees.
To examine if this makes a difference, we consider a ``skip-sequence'' task that samples subsequences of the list of tokens instead of sampling subexpressions.
We generated three datasets for the skip-sequence task, where we sample subsequences of different lengths (short/medium/long).
For the task ``skip-sequence (long)'', we pick two positions in the token sequence at uniformly at random and select the sequence that is between them.
For the tasks ``skip-sequence (medium)'' and ``skip-sequence (short)'', we limit their distance to 100 and 50 tokens, respectively.

% \paragraph{Training dataset statistics}
% In Table~\ref{tab:datastats} we give a compact overview of the basic statistics, such as the size of the training sets and the average length of the sequences.

\begin{table}[h]
    \centering
    \begin{tabular}{|l||c|c|c|}
    \hline
        Dataset & \# examples & \# tokens (input/output) & avg length (input/output) \\\hline\hline
        Skip-tree (weighted) & 25.8M & 17.4B/1.6B & 675/61 \\
        Skip-tree (uniform) & 25.7M & 18.8B/316M & 732/12 \\
        
        % {"examples_test": 7299456, "examples_train": 25821877, "examples_unknown": 12496, "examples_valid": 7356537, "lines read": 21053, "proof logs": 21053, "input_tokens": 29007199800, "output_tokens": 2643848199}
        
        Skip-tree (small) & 5.2M & 3.5B/521M & 673/100 \\
        % Skip-tree (tiny) & & &  \\
        Skip-tree (no <MASK>) & 25.8M & 19.4B/1.6B & 750/61 \\
        Skip-sequence (long) & 19.2M & 11.9B/2.8B & 620/146 \\
        Skip-sequence (medium) & 26.0M & 19.4B/884M & 744/34 \\
        Skip-sequence (short) & 26.0M & 19.6B/479M & 752/18 \\
        \hline
    \end{tabular}
    \vspace{2mm}
    \caption{Basic statistics of the \emph{training splits} of the data sets. Number of tokens in the training set measured before padding.}
    \label{tab:datastats}
    \vspace{-2mm}
\end{table}

%%%%%%%%%%%%%%%%%%%%%%%%%%%%%%%%%%%%%%%%%%%%%%%%%%%%%%%%%%%%%%%%%%%%%%%%
\section{Evaluation Tasks}
\label{sec:evaluationtasks}
%%%%%%%%%%%%%%%%%%%%%%%%%%%%%%%%%%%%%%%%%%%%%%%%%%%%%%%%%%%%%%%%%%%%%%%%

In this section we suggest several logical reasoning tasks on which our language models can be evaluated.
These tasks require different levels of logical reasoning, ranging from mostly mechanical application of typing rules to conjecturing under which assumptions a statement might hold.

We intentionally define them to be out-of-distribution
% to various degrees
compared to the training data.
Not only do we generate the examples in a slightly different way, we also generate them from the validation set of the theorem database.
That is, the model has never seen the source data, nor has it seen the proofs of these theorems.
This makes the tasks more challenging, and also ensures that we force the models to go beyond memorization.
To give the interested reader a better impression of the evaluation tasks, we provide a list of randomly selected examples in Appendix~\ref{app:randomexamples}.

\paragraph{Type Inference.}
We generate type inference problems similar to how we generated the skip-tree training data, which we described in Section~\ref{sec:skiptree}.
However, we restrict the sampling of subexpressions to subtrees that represent types of variables or constants (i.e. not fragments of other types).
% Note that during sampling, the weighting by the number of tokens of the subtrees still applies.

We generated two variants of the type inference task:
In the task we call ``Type Inference,'' we replace only the selected type by the <PREDICT> token and do not mask out anything else.
In the second variant we name ``Hard Type Inference,'' we additionally replace \emph{all other types} by the <MASK> token.
The two tasks loosely correspond to the deriving the first and the last type during type inference.

For example, consider $x = x$, which in the s-expression syntax is represented as follows:
\[
\texttt{(a (a (c (fun (A) (fun (A) (bool))) =) (v A x)) (v A x))}
\]
Each subexpression here is either a leaf or a triple.
The first element of these triples indicates their kind: \texttt{a} indicates function applications, \texttt{c} indicates constants (i.e. symbols that have been defined in the formal system), \texttt{v} indicates a variable, and finally \texttt{fun} indicates a function type.
The equality operator ``\texttt{=}'' is represented by \texttt{(c (fun (A) (fun (A) (bool))) =)}, which indicates that it is a constant that has a function type taking two arguments of arbitrary type \texttt{A} and returns a bool.
Since functions are typically curried in this representation, we have two function applications, both times with the variable \texttt{x} as the argument.

An example for the ``Type Inference'' evaluation task would be:
\[
\texttt{(a (a (c <PREDICT> =) (v A x)) (v A x))}
\]
The type of the equality operator is still uniquely defined, as we know what the equality is applied to (two arguments of type \texttt{A}) and because top-level application always has to return a boolean value.
In this example the type could have been computed by a classical type inference algorithm.

For the ``Hard Type Inference'' evaluation task, the input would look as follows:
\[
\texttt{(a (a (c <PREDICT> =) (v <MASK> x)) (v <MASK> x))}
\]
Now, the type inference task is highly ambiguous.
In fact, in this case, variable \texttt{x} could have any type, and the equality operator would have to adapt to the type of its arguments accordingly.
Further, note that the hard type inference task masks out many more subtrees compared to the training data.

\paragraph{Assumptions.}
This evaluation task is to predict missing assumptions for theorems in the validation set.
We extract these tasks by searching for ``top-level implications'' and replacing their left operand by the <PREDICT> token. 
We define an implication operator ``$\Rightarrow$'' in an expression to be a \emph{top-level implication} if it is either the top-most operator of the expression, or occurs only under quantifiers, conjunctions, disjunctions, or on the right side of other top-level implications.
This definition helps us to avoid picking assumptions in negated parts of formulas.

Note that we can have multiple top-level implications per validation theorem.
Consider the abstracted example $(a\Rightarrow b) \wedge (c \Rightarrow (d \Rightarrow e))$.
In this case, $a$, $c$, and $d$ are all considered to be assumptions of top-level implications.

An example from the theorem database is $x=y \Rightarrow a + x = a + y$, for which the task is to predict $x=y$ given <PREDICT> $\Rightarrow a + x = a + y$. 
(We omit the presentation of this example as an s-expression for the sake of readability.)
At first, the expression to predict in this case may seem unique, but there are actually many ways to complete the task into a true statement; e.g. $y=x$ or $x=0\wedge y=0$.
Still, most humans would likely guess $x=y$ as it is simple and general, and because $x$ occurs before $y$ in the alphabet.
To make a correct prediction, our language models thus have to understand which statements are more general and also know about naming conventions.

Below we give some examples of this reasoning task that we selected for their simplicity.
(For a representative selection, see Appendix~\ref{app:randomexamples}.)
While it is often easy to ``see'' that a given solution to such a task is correct, it can be non-trivial to come up with a solution in the first place.
We encourage the reader to make their own predictions before looking up the ground truth in Appendix~\ref{app:exampletasks}:
\begin{itemize}
    \item $\texttt{<PREDICT>}  \Rightarrow (x \Leftrightarrow (~b \vee x1) \wedge (b \vee x0))$
    \item $\texttt{<PREDICT>} \Rightarrow (g \setminus \{s\}) = g$
    \item $\texttt{<PREDICT>} \Rightarrow (x1 / y1 = x2 / y2 ~\Leftrightarrow~ x1 * y2 = x2 * y1)$
\end{itemize}

\paragraph{Equalities.}
Similar to the task of predicting missing assumptions, we ask to predict one side of a top-level equality in this task.
Again, we define top-level equalities to be any equality that occurs as the top-level operator of the formula or occurs inside quantifiers, conjunctions, disjunctions, or on the right side of implications.
For example, from the theorem $\forall x. x = (x = \texttt{True})$ we extract two evaluation examples: $\forall x.~ \texttt{<PREDICT>} = (x = \texttt{True})$ and $\forall x.~ x = \texttt{<PREDICT>}$.

Again, we present some simple example tasks (in mathematical notation for the sake of readability) and provide the ground truth as well as the model predictions in Appendix~\ref{app:exampletasks}:
\begin{itemize}
    \item $\forall x, n\in\mathbb{N}:~ (x^n = 1) = \texttt{<PREDICT>}$
    \item $\forall m, n:~ n \leq m \Rightarrow m - n + n = \texttt{<PREDICT>}$
    \item $\forall l, m:~ \texttt{<PREDICT>} = \texttt{APPEND}(\texttt{REVERSE}(m),~ \texttt{REVERSE}(l))$
\end{itemize}

\section{Results and Discussion}
\label{sec:results}

We trained a Transformer with the hyperparameters specified in the appendix on the skip-tree dataset and each of the ablations for 1M steps with a batch size of 256.
% We first compare and discuss the ability of the trained models to predict the ground truth.

In language modeling for natural language one of the key metrics is how often the next token in the ground truth is correctly predicted.
This is not an ideal measurement for formal mathematics as even a single incorrect token can invalidate the entire statement. %, while in natural language such mistakes are sometimes tolerable.
Also, the s-expression representation is relatively lengthy and barely human-readable, so a token-level measurement does not allow us to compare our models to the natural language models in any case.
\emph{In the first part of our evaluation we therefore focus on exact matches of the entire predicted statement}.

\begin{table}[h]
    \centering
    \begin{tabular}{|l||c|c|c|c|}
    \hline
        Dataset & Type Inference & Hard Type Inference & Assumptions & Equalities \\\hline\hline
        Skip-tree (uniform) & 96.21\% & \bf 94.40\% & 40.85\% & \bf 46.57\% \\
        Skip-tree (weighted) & \bf 96.23\% & 93.32\% & \bf 40.86\% & 42.89\% \\
        Skip-tree (small) & 95.89\% & 90.42\% & 39.23\% & 40.91\% \\
        % Skip-tree (tiny) & & & &  \\
        Skip-tree (no <MASK>) & 96.07\% & {\color{gray}32.50\%} & 38.38\% & 41.60\% \\
        Skip-sequence (long) & 9.44\% & {\color{gray}0.06\%} & 0.53\% & 0.56\% \\
        Skip-sequence (medium) & 48.94\% & {\color{gray}5.97\%} & 3.32\% & 3.55\% \\
        Skip-sequence (short) & 77.25\%\% & {\color{gray}3.21\%} & 0.68\% & 2.06\% \\
        \hline
    \end{tabular}
    \vspace{2mm}
    \caption{Success rate of predicting the ground truth in a beam search of width 8 after training a model on various datasets.
    Grayed out values indicate experiments where the training data did not include the <MASK> token but the evaluation data did.}
    \label{tab:exact_beam8}
    \vspace{-3mm}
\end{table}

In Table~\ref{tab:exact_beam8} we present how well the Transformer model, trained on different datasets, can predict the ground truth sequences.
We can observe that for type inference, i.e. the more mechanical reasoning tasks, the models achieve a high accuracy - even in the Hard Type Inference case where the expression was stripped of all types.
We see that the skip-tree task and its ablations clearly dominate the skip-sequence language modeling task.
There does not seem to be a major difference between the ``uniform'' and ``weighted'' sampling strategies for the skip-tree model.

A closer inspection of the skip-sequence model shows that its predictions rarely parse or typecheck.
On manual inspection of the predictions, it seems that the skip-sequence models consistently add surplus tokens at the end, or stop expressions too early; \emph{they appear to be unable to correctly identify the end of the expression to predict}.

\subsection{Conjecturing}

In the experiments above, we measured how often the models predicted the ground truth in the evaluation tasks.
We now change our point of view, and examine whether the models can be used to generate new conjectures.
We define conjectures as mathematical statements that \emph{differ from the ground truth and any expression the model has seen during training}. 
Additionally, a meaningful conjecture should be syntactically correct, typecheck, be provable, and be useful in the context of other proofs.

Since the training data is derived exclusively from true statements (i.e. human proof steps), the language models are incentivized to complete partial statements in a way that makes them true.
Presented with one of the evaluation tasks, to predict missing assumptions or to predict the missing side of an equation, the models may thus complete these statements in multiple ways that make them true.
The predictions that do not match the ground truth may still be true and useful statements.
In the following we describe experiments that help us estimate how often this is the case.

\paragraph{Free-form conjecturing.}
In addition to the ``assumptions'' and the ``equalities'' evaluation tasks, we consider a third task for producing conjectures.
In this task, which we call ``free-form conjecturing'', we query the model with a single prompt: \texttt{(\texttt{<theorem>} \texttt{<PREDICT>})}.
This helps us to analyze what the language models produce when given no context.
The \texttt{<theorem>} tag indicates only that the statement should be a theorem, and not an intermediate proof step, which would start with the \texttt{<goal>} tag.
For free-form conjecturing we want to produce a variety of different predictions, and thus use a beam search with high beam width of 1024.
We did not include the free-form conjecturing task in Table~\ref{tab:exact_beam8}, as there is no ground truth to match against.

\paragraph{How often are predictions true and new?} For this measurement, we replace the <PREDICT> token with the predicted sequence and attempt to prove the resulting statement in the DeepHOL theorem prover~\citep{bansal2019holist}.
Note that this can only give us a \emph{lower bound to the number of true statements}, because of the limitations of the prover: The version of the DeepHOL theorem prover used here can prove around 58\% of the validation theorems.
So we expect the estimates here to be considerably below the number of actually true statements.

In Table~\ref{tab:truth_beam8} we report two numbers for each evaluation task:
The first number is the percentage of generated statements known to be provable, including exact matches, statements from the training set, and statements provable with DeepHOL.
The second number is the percentage of generated statements that are provable and \emph{new} - excluding exact matches with the ground truth and statements from the training set.
The denominator for both numbers is the same: the set of all predictions from the beam searches in Table~\ref{tab:exact_beam8}.

% We want to exclude statements that the model might have memorized, and thus filter out all exact matches with the training set.
% Matches with statements in the training set can happen if there are two similar statements in the training set and the validation set, and the only subexpression by which they differ has been replaced by the <PREDICT> token.
% Additionally, we filter out the exact matches with the ground truth, such that we measure only the number of new statements that are provable. 

% In Table~\ref{tab:truth_beam8} we present how many of these conjectures generated by the model were provable.
We believe that these measurements show a significant bias towards true statements.
While in some tasks, less than half of the statements were provable, there are simply many more ways to write a false statement than a true statement.

\begin{table}[h]
    \centering
    \begin{tabular}{|l||c|c|c|c|}
    \hline
        Dataset & Assumptions & Equalities & Free-form Conjecturing \\\hline\hline
        Skip-tree (uniform) & 32.19\%/26.20\% & 19.61\%/12.28\% & 82.32\%/12.70\% \\
        Skip-tree (weighted) & 32.41\%/26.91\% & 17.96\%/11.63\% & 97.75\%/0.59\% \\
        % Skip-tree (medium) & TODO\% & TODO\% & - \\
        % Skip-tree (no <MASK>) & TODO\% & TODO\% & - \\
        % Skip-sequence (long) & TODO\% & TODO\% & -\\
        \hline
    \end{tabular}
    \vspace{2mm}
    \caption{Percentage of ``provable statements''/``provable {\bf new} statements''.
    % The ``true statements'' include the provable statements, matches with the ground truth, and matches with any known true statement from our training data.
    % The provable new statements only include statements that did \emph{not} match the ground truth or any statement from the training data, and that were provable by the theorem prover.
    The type inference tasks are not included as we are only interested in the predictions that do not match the ground truth. For the type inference tasks, these statements are either semantically equivalent to existing statements or statements that do not type check.
    }
    \label{tab:truth_beam8}
    \vspace{-4mm}
\end{table}

\paragraph{Are the conjectures useful?} For some evaluation tasks, the models could ``cheat'' on the truth metric by making the statements \emph{trivially} true.
For example, the models can predict \texttt{False} as an assumption, or complete the missing part of an equation by making it an identity (e.g. complete \texttt{x = <PREDICT>} by predicting \texttt{x}).
In fact, manual inspection revealed several such cases.

To make this measurable, we added the provable statements to the theorem database, and ran the reinforcement learning experiments of the DeepHOL theorem prover~\citep{bansal2019holist} to measure how many of the statements were used as premises.
In this experiment we also make sure that the new theorems cannot be used in the proofs of their premises.
% In Table~\ref{tab:usefulness_beam8} we report the number of statements that the system learns to use as premises.
In a ``pruning'' step DeepHOL minimizes proofs by removing each individual premise in a proof and checking if the proof still holds.
Only the premises that survive this step are classified as \emph{useful}.
While this measurement is a relatively low bar, it filters out statements that have no effect in any proof.

We ran three reinforcement learning experiments, one for each of the evaluation tasks.
We then measured how many of the theorems generated by each task are used as a premise in one of the over 200,000 proofs found for each of the experiments.
For the assumptions task, 3445 of the 3857 theorems were used at least once.
For the equalities task and the free-form conjectures it was 979 out of 3440 and 49 out of 130, respectively.
We provide usage histograms in Appendix~\ref{app:usage}.

While some of the most frequently used conjectures turned out to be alpha-equivalent variations of existing theorems in the theorem database, we found some interesting examples among the most used conjectures:
% Some of the generated theorems were used frequently.
% The three most used new theorems are:
\begin{itemize}
    \item Assumptions task, 1728 usages: $b = a + c \Rightarrow a = b - c$.
    Humans have used this theorem over vector arithmetic in many proofs. However, this theorem has always been defined as a \emph{local} lemma and thus did not made it into the theorem database.
    This conjecture apparently filled a gap in the theorem database.
    \item Free-form conjecturing task, 15 usages: $\texttt{COUNTABLE}(\{s(n) \mid n \in \mathbb{N}\})$.
    In contrast to the previous example, there are no occurrences of this statement (or an equivalent statement) in the theorem database or any human proof, not even as a local lemma.
\end{itemize}

% While the theorems number 2 and 3 in the list above are alpha-equivalent to theorems that already existed in our theorem database, the first theorem revealed something interesting:

% Only at a very late point in the theorem database a variant of this theorem occurs.

% Table~\ref{tab:usefulness_beam8} shows that a significant portion of the provable statements turned out to be useful in some other proof.
% Of course, this is after several filtering steps and makes only a small number of the overall predictions.
These results suggest that self-supervised language models show some ability to produce new, useful conjectures, even without fine tuning or specialized training.
% To effectively produce conjectures that are useful in a specific context, we expect that a more targeted training approach is needed.
% This is after filtering out matches with the ground truth and any statement that we trained the models, and any statements that the theorem prover was unable to prove.

%%%%%%%%%%%%%%%%%%%%%%%%%%%%%%%%%%%%%%%%%%%%%%%%%%%%%%%%%%%%%%%%%%%%%%%%
\section{Conclusion}
\label{sec:conclusion}
%%%%%%%%%%%%%%%%%%%%%%%%%%%%%%%%%%%%%%%%%%%%%%%%%%%%%%%%%%%%%%%%%%%%%%%%

In this work, we applied the paradigms of self-supervised language modeling to formal mathematics and show that this leads to surprisingly good reasoning capabilities.
We introduced a novel self-supervised training task for formal mathematics that outperforms existing training tasks used for natural language.
We also suggested several evaluation tasks for measuring mathematical reasoning capabilities of language models for formal mathematics without the need of fine tuning.
% Our experiments demonstrate that language models are already surprisingly capable at a variety of reasoning tasks that they were not trained for directly.
Finally, we explored the ability of language models to produce new conjectures by measuring how many of the new predictions are provable and useful for other proofs.

% Besides making progress towards measuring mathematical reasoning abilities,
% our experiments suggest that the language modeling approach also works for formal languages.
% While we apply the approach only to formal mathematics, it could just as well be applied to programming languages.
% This could lead to strong generative models for both mathematics and code (i.e. program synthesis), and also be used as pretraining task for these domains.

%%%%%%%%%%%%%%%%%%%%%%%%%%%%%%%%%%%%%%%%%%%%%%%%%%%%%%%%%%%%%%%%%%%%%%%%
\section*{Broader Impact}
%%%%%%%%%%%%%%%%%%%%%%%%%%%%%%%%%%%%%%%%%%%%%%%%%%%%%%%%%%%%%%%%%%%%%%%%

Our ambition is to create strong automated reasoning systems.
In the long run, such systems could be used as a tool in mathematical research, engineering, and other sciences.
Such systems could be used as stand alone tools, but also as a component of other systems that utilizes mathematical reasoning, such as in verification of software and hardware systems and physical modeling and exploration.
This should be very helpful for accelerating scientific progress.

In its current form, however, the methods presented in the paper are not applicable directly to solving any particular scientific tasks.
Therefore we do not anticipate any ethical or fairness issues arising from direct applications of the technologies presented here.
However if our methods are trained for mimicking human reasoning in domains that argue over, for example, personal data, then it might reinforce human biases present in the dataset.
On the other hand, the abstraction capabilities of our system might make those biases more explicit and interpretable and could help exposing them.
As part of a larger software verification system, some of the methods presented here might be used for automated reverse engineering the internal working of other software systems, finding and exploiting vulnerabilities in them.

\bibliography{main}

\newpage
\appendix

\section{Hyperparameters}

We trained the Transformers with these hyperparameters:

\begin{itemize}
    \item vocabulary size: 1200
    \item embedding size: 128
    \item attention dropout: 0.1
    \item nonlinearity: gelu
    \item hidden layer dropout: 0.1
    \item hidden layer size: 512
    \item initializer range: 0.02
    \item intermediate size: 768
    \item number of attention heads: 8
    \item number of hidden layers in encoder: 2
    \item number of hidden layers in decoder: 4
\end{itemize}

\section{Usage Statistics of Conjectures}
\label{app:usage}

\begin{figure}[h]
    \centering
    \includegraphics[width=.3\textwidth]{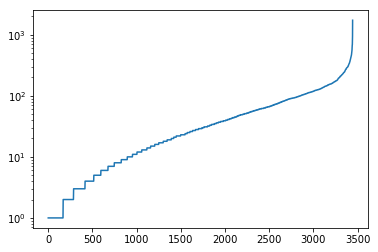}
    \includegraphics[width=.3\textwidth]{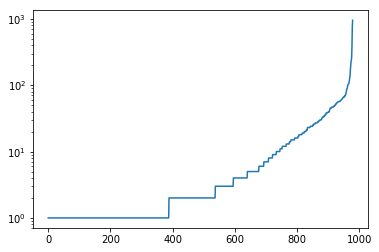}
    \includegraphics[width=.3\textwidth]{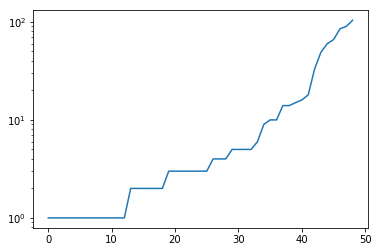}
    \caption{Histograms of premise usage of the conjectures generated through the assumptions task (left), the equality task (middle), and through free-form conjecturing (right).
    X-axes are the new theorems, sorted by number of usages.
    Y-axes indicate the number of usages on a log scale.}
    \label{fig:my_label}
\end{figure}

\section{A Close Look at Simple Example Tasks}
\label{app:exampletasks}

\paragraph{Assumptions.}
In Section~\ref{sec:evaluationtasks} we presented the following three examples of the task to predict missing assumptions. For the sake of readability we here discuss only the pretty printed versions. For examples in s-expression syntax, please visit Appendix~\ref{app:randomexamples}.
\begin{itemize}
    \item $\texttt{<PREDICT>}  \Rightarrow (x \Leftrightarrow (~b \vee x1) \wedge (b \vee x0))$
    \item $\texttt{<PREDICT>} \Rightarrow (g \setminus \{s\}) = g$
    \item $\texttt{<PREDICT>} \Rightarrow (x1 / y1 = x2 / y2 ~\Leftrightarrow~ x1 * y2 = x2 * y1)$
\end{itemize}

The ground truth answers are as follows:
\begin{itemize}
    \item $((b \Leftrightarrow \texttt{False}) \Rightarrow (x \Leftrightarrow x0)) \wedge (b\Leftrightarrow \texttt{True}) \Rightarrow (x \Leftrightarrow x1)$
    \item $\neg (s \in g)$
    \item $0 < y1 \wedge 0 < y2$, note that $0 \neq y1 \wedge 0 \neq y2$ would be a more general assumption.
\end{itemize}

For the first and the third task, the language model ``skip-tree (weighted)'' makes a correct prediction in the top 3 candidates in a beam search of width 8.
For the seconds task, the language model mostly produces incorrectly typed expressions: it appears to think that $s$ is a set of the same type as $g$.

\paragraph{Equalities.} We presented these examples for the equality evaluation task:
\begin{itemize}
    \item $\forall x, n\in\mathbb{N}:~ (x^n = 1) = \texttt{<PREDICT>}$
    \item $\forall m, n:~ n \leq m \Rightarrow m - n + n = \texttt{<PREDICT>}$
    \item $\forall l, m:~ \texttt{<PREDICT>} = \texttt{APPEND}(\texttt{REVERSE}(m),~ \texttt{REVERSE}(l))$
\end{itemize}

The ground truth for the tasks is:
\begin{itemize}
    \item $x = 1 \vee n = 0$
    \item $m$
    \item $\texttt{REVERSE}(\texttt{APPEND}(l, m))$
\end{itemize}

Examples two and three are predicted correctly in a beam search with beam width 8.
For the first example, the model almost gets it correct in two of the 8 attempts: $x = 1 \vee n = 1$, and $x = 0 \vee n = 1$.
We find it surprising that the model apparently understands that there are two cases to consider, but that the exact combination of constants (1 and 0) is a challenge.

% "(<theorem> (a (c (fun (fun (num) (bool)) (bool)) !) (l (v (num) x) (a (c (fun (fun (num) (bool)) (bool)) !) (l (v (num) n) (a (a (c (fun (bool) (fun (bool) (bool))) =) (a (a (c (fun (num) (fun (num) (bool))) =) (a (a (c (fun (num) (fun (num) (num))) EXP) (v (num) x)) (v (num) n))) (a (c (fun (num) (num)) NUMERAL) (a (c (fun (num) (num)) BIT1) (c (num) _0))))) <TO_PREDICT>))))))"
% "<START> (a (a (c (fun (bool) (fun (bool) (bool))) \\/) (a (a (c (fun (num) (fun (num) (bool))) =) (v (num) x)) (a (c (fun (num) (num)) NUMERAL) (a (c (fun (num) (num)) BIT1) (c (num) _0))))) (a (a (c (fun (num) (fun (num) (bool))) =) (v (num) n)) (a (c (fun (num) (num)) NUMERAL) (c (num) _0)))) <END>"

\section{Randomly Selected Example Tasks}
\label{app:randomexamples}

In the following, we provide a list of 5 examples for each of the evaluation tasks, sampled uniformly at random.
% For the type inference tasks, we present the type inference tasks in the actual s-expression syntax, but for the other tasks, we resort to the easier-to-read HOL Light syntax.

\paragraph{Type Inference.}
\begin{itemize}
\item \texttt{(<theorem> (a (c <PREDICT> !) (l (v (fun (cart (real) ?1) (bool)) t) (a (c (fun (fun (fun (cart (real) ?1) (bool)) (bool)) (bool)) !) (l (v (fun (cart (real) ?1) (bool)) u) (a (a (c (fun (bool) (fun (bool) (bool))) ==>) (a (a (c (fun (bool) (fun (bool) (bool))) $\wedge$) (a (c (fun (fun (cart (real) ?1) (bool)) (bool)) !) (l (v (cart (real) ?1) b) (a (a (c (fun (bool) (fun (bool) (bool))) $\vee$) (a (c (fun (fun (cart (real) ?1) (bool)) (bool)) ?) (l (v (cart (real) ?1) w) (a (a (c (fun (bool) (fun (bool) (bool))) $\wedge$) (a (a (c (fun (cart (real) ?1) (fun (fun (cart (real) ?1) (bool)) (bool))) IN) (v (cart (real) ?1) w)) (v (fun (cart (real) ?1) (bool)) t))) (a (a (c (fun (cart (real) ?1) (fun (fun (cart (real) ?1) (bool)) (bool))) IN) (v (cart (real) ?1) w)) (a (c (fun (prod (cart (real) ?1) (real)) (fun (cart (real) ?1) (bool))) ball) (a (a (c (fun (cart (real) ?1) (fun (real) (prod (cart (real) ?1) (real)))) ,) (v (cart (real) ?1) b)) (a (c (fun (num) (real)) real\_of\_num) (a (c (fun (num) (num)) NUMERAL) (a (c (fun (num) (num)) BIT1) (c (num) \_0))))))))))) (a (c (fun (fun (cart (real) ?1) (bool)) (bool)) ?) (l (v (cart (real) ?1) w) (a (a (c (fun (bool) (fun (bool) (bool))) $\wedge$) (a (a (c (fun (cart (real) ?1) (fun (fun (cart (real) ?1) (bool)) (bool))) IN) (v (cart (real) ?1) w)) (v (fun (cart (real) ?1) (bool)) u))) (a (a (c (fun (cart (real) ?1) (fun (fun (cart (real) ?1) (bool)) (bool))) IN) (v (cart (real) ?1) w)) (a (c (fun (prod (cart (real) ?1) (real)) (fun (cart (real) ?1) (bool))) ball) (a (a (c (fun (cart (real) ?1) (fun (real) (prod (cart (real) ?1) (real)))) ,) (v (cart (real) ?1) b)) (a (c (fun (num) (real)) real\_of\_num) (a (c (fun (num) (num)) NUMERAL) (a (c (fun (num) (num)) BIT1) (c (num) \_0)))))))))))))) (a (c (fun (fun ?0 (bool)) (bool)) !) (l (v ?0 x) (a (a (c (fun (bool) (fun (bool) (bool))) ==>) (a (a (c (fun ?0 (fun (fun ?0 (bool)) (bool))) IN) (v ?0 x)) (v (fun ?0 (bool)) d))) (a (c (fun (bool) (bool)) $\sim$) (a (a (c (fun (cart (real) ?1) (fun (fun (cart (real) ?1) (bool)) (bool))) IN) (a (v (fun ?0 (cart (real) ?1)) g) (v ?0 x))) (a (a (c (fun (fun (cart (real) ?1) (bool)) (fun (fun (cart (real) ?1) (bool)) (fun (cart (real) ?1) (bool)))) UNION) (v (fun (cart (real) ?1) (bool)) t)) (v (fun (cart (real) ?1) (bool)) u))))))))) (a (c (fun (bool) (bool)) $\sim$) (a (c (fun (fun (cart (real) ?1) (bool)) (bool)) ?) (l (v (cart (real) ?1) b) (a (a (c (fun (fun (cart (real) ?1) (bool)) (fun (fun (cart (real) ?1) (bool)) (bool))) SUBSET) (a (c (fun (prod (cart (real) ?1) (real)) (fun (cart (real) ?1) (bool))) ball) (a (a (c (fun (cart (real) ?1) (fun (real) (prod (cart (real) ?1) (real)))) ,) (v (cart (real) ?1) b)) (a (c (fun (num) (real)) real\_of\_num) (a (c (fun (num) (num)) NUMERAL) (a (c (fun (num) (num)) BIT1) (c (num) \_0))))))) (a (a (c (fun (fun ?0 (cart (real) ?1)) (fun (fun ?0 (bool)) (fun (cart (real) ?1) (bool)))) IMAGE) (v (fun ?0 (cart (real) ?1)) g)) (v (fun ?0 (bool)) d))))))))))))}

Ground truth: \texttt{<START> (fun (fun (fun (cart (real) ?1) (bool)) (bool)) (bool)) <END>}

\item \texttt{(<theorem> (a (c <PREDICT> !) (l (v (fun (cart (real) N) (bool)) s) (a (a (c (fun (bool) (fun (bool) (bool))) =) (a (c (fun (fun (cart (real) N) (bool)) (bool)) is\_interval) (a (a (c (fun (fun (cart (real) N) (cart (real) N)) (fun (fun (cart (real) N) (bool)) (fun (cart (real) N) (bool)))) IMAGE) (c (fun (cart (real) N) (cart (real) N)) vector\_neg)) (v (fun (cart (real) N) (bool)) s)))) (a (c (fun (fun (cart (real) N) (bool)) (bool)) is\_interval) (v (fun (cart (real) N) (bool)) s))))))}

Ground truth: \texttt{<START> (fun (fun (fun (cart (real) N) (bool)) (bool)) (bool)) <END>}

\item \texttt{(<theorem> (a (c (fun (fun (real) (bool)) (bool)) !) (l (v (real) x) (a (a (a (c (fun (fun (real) (real)) (fun (real) (fun (net (real)) (bool)))) has\_real\_derivative) (c (fun (real) (real)) atn)) (a (c (fun (real) (real)) real\_inv) (a (a (c (fun (real) (fun (real) (real))) real\_add) (a (c (fun (num) (real)) real\_of\_num) (a (c (fun (num) (num)) NUMERAL) (a (c (fun (num) (num)) BIT1) (c (num) \_0))))) (a (a (c (fun (real) (fun (num) (real))) real\_pow) (v <PREDICT> x)) (a (c (fun (num) (num)) NUMERAL) (a (c (fun (num) (num)) BIT0) (a (c (fun (num) (num)) BIT1) (c (num) \_0)))))))) (a (c (fun (real) (net (real))) atreal) (v (real) x))))))}

Ground truth: \texttt{<START> (real) <END>}

\item \texttt{(<theorem> (a (a (c (fun (fun ?0 (bool)) (fun (fun ?0 (bool)) (bool))) =) (a (a (c (fun (fun ?0 (bool)) (fun (fun ?0 (bool)) (fun ?0 (bool)))) INTER) (v (fun ?0 (bool)) s)) (a (a (c (fun (fun ?0 (bool)) (fun (fun ?0 (bool)) (fun ?0 (bool)))) UNION) (v (fun ?0 (bool)) t)) (v (fun ?0 (bool)) u)))) (a (a (c (fun (fun ?0 (bool)) (fun (fun ?0 (bool)) (fun ?0 (bool)))) UNION) (a (a (c <PREDICT> INTER) (v (fun ?0 (bool)) s)) (v (fun ?0 (bool)) t))) (a (a (c (fun (fun ?0 (bool)) (fun (fun ?0 (bool)) (fun ?0 (bool)))) INTER) (v (fun ?0 (bool)) s)) (v (fun ?0 (bool)) u)))))}

Ground truth: \texttt{<START> (fun (fun ?0 (bool)) (fun (fun ?0 (bool)) (fun ?0 (bool)))) <END>}

\item \texttt{(<theorem> (a (a (c (fun (real) (fun (real) (bool))) =) (a (c (fun (cart (real) ?0) (real)) infnorm) (a (c (fun (num) (cart (real) ?0)) vec) (a (c (fun (num) (num)) NUMERAL) (c (num) \_0))))) (a (c (fun (num) (real)) real\_of\_num) (a (c (fun (num) (num)) NUMERAL) (c <PREDICT> \_0)))))}

Ground truth: \texttt{<START> (num) <END>}
\end{itemize}

\paragraph{Hard Type Inference.}
\begin{itemize}
\item \texttt{(<theorem> (a (c <MASK> !) (l (v <MASK> s) (a (a (c <MASK> =) (a (c <MASK> INTERS) (v <MASK> s))) (a (a (c <PREDICT> DIFF) (c <MASK> UNIV)) (a (c <MASK> UNIONS) (a (c <MASK> GSPEC) (l (v <MASK> GEN\%PVAR\%0) (a (c <MASK> ?) (l (v <MASK> t) (a (a (a (c <MASK> SETSPEC) (v <MASK> GEN\%PVAR\%0)) (a (a (c <MASK> IN) (v <MASK> t)) (v <MASK> s))) (a (a (c <MASK> DIFF) (c <MASK> UNIV)) (v <MASK> t)))))))))))))}

Ground truth: \texttt{<START> (fun (fun ?0 (bool)) (fun (fun ?0 (bool)) (fun ?0 (bool)))) <END>}

\item \texttt{(<theorem> (a (c <MASK> !) (l (v <MASK> f) (a (c <MASK> !) (l (v <MASK> s) (a (a (c <MASK> =) (a (a (c <MASK> uniformly\_continuous\_on) (v <MASK> f)) (v <MASK> s))) (a (c <MASK> !) (l (v <MASK> e) (a (a (c <MASK> ==>) (a (a (c <MASK> real\_lt) (a (c <MASK> real\_of\_num) (a (c <MASK> NUMERAL) (c <MASK> \_0)))) (v <MASK> e))) (a (c <MASK> ?) (l (v <MASK> d) (a (a (c <MASK> $\wedge$) (a (a (c <MASK> real\_lt) (a (c <MASK> real\_of\_num) (a (c <MASK> NUMERAL) (c <MASK> \_0)))) (v <MASK> d))) (a (c <MASK> !) (l (v <MASK> t) (a (c <MASK> !) (l (v <MASK> t') (a (a (c <MASK> ==>) (a (a (c <MASK> $\wedge$) (a (a (c <MASK> SUBSET) (v <MASK> t)) (v <MASK> s))) (a (a (c <MASK> $\wedge$) (a (a (c <MASK> SUBSET) (v <PREDICT> t')) (v <MASK> s))) (a (a (c <MASK> $\wedge$) (a (c <MASK> bounded) (v <MASK> t))) (a (a (c <MASK> $\wedge$) (a (c <MASK> bounded) (v <MASK> t'))) (a (a (c <MASK> real\_lt) (a (c <MASK> hausdist) (a (a (c <MASK> ,) (v <MASK> t')) (v <MASK> t)))) (v <MASK> d))))))) (a (a (c <MASK> real\_lt) (a (c <MASK> hausdist) (a (a (c <MASK> ,) (a (a (c <MASK> IMAGE) (v <MASK> f)) (v <MASK> t'))) (a (a (c <MASK> IMAGE) (v <MASK> f)) (v <MASK> t))))) (v <MASK> e)))))))))))))))))))}

Ground truth: \texttt{<START> (fun (cart (real) M) (bool)) <END>}

\item \texttt{(<theorem> (a (a (c <MASK> ==>) (a (a (c <PREDICT> IN) (v <MASK> a)) (v <MASK> s))) (a (a (c <MASK> =) (a (a (c <MASK> DIFF) (a (a (c <MASK> INSERT) (v <MASK> a)) (a (a (c <MASK> DELETE) (v <MASK> t)) (v <MASK> b)))) (v <MASK> s))) (a (a (c <MASK> DELETE) (a (a (c <MASK> DIFF) (v <MASK> t)) (v <MASK> s))) (v <MASK> b)))))}

Ground truth: \texttt{<START> (fun ?0 (fun (fun ?0 (bool)) (bool))) <END>}

\item \texttt{(<theorem> (a (c <MASK> !) (l (v <PREDICT> b) (a (c <MASK> convex) (a (c <MASK> GSPEC) (l (v <MASK> GEN\%PVAR\%0) (a (c <MASK> ?) (l (v <MASK> z) (a (a (a (c <MASK> SETSPEC) (v <MASK> GEN\%PVAR\%0)) (a (a (c <MASK> real\_gt) (a (c <MASK> Im) (v <MASK> z))) (v <MASK> b))) (v <MASK> z))))))))))}

Ground truth: \texttt{<START> (real) <END>}

\item \texttt{(<theorem> (a (c <MASK> !) (l (v <MASK> x) (a (a (c <MASK> ==>) (a (c <MASK> $\sim$) (a (a (c <MASK> nadd\_eq) (v <MASK> x)) (a (c <MASK> nadd\_of\_num) (a (c <MASK> NUMERAL) (c <MASK> \_0)))))) (a (c <MASK> ?) (l (v <MASK> B) (a (c <MASK> ?) (l (v <MASK> N) (a (c <MASK> !) (l (v <MASK> m) (a (c <MASK> !) (l (v <MASK> n) (a (a (c <MASK> ==>) (a (a (c <MASK> $\wedge$) (a (a (c <MASK> <=) (v <MASK> N)) (v <MASK> m))) (a (a (c <MASK> <=) (v <MASK> N)) (v <MASK> n)))) (a (a (c <MASK> <=) (a (a (c <MASK> *) (a (a (c <MASK> *) (a (a (c <MASK> dest\_nadd) (v <MASK> x)) (v <MASK> m))) (a (a (c <MASK> dest\_nadd) (v <MASK> x)) (v <MASK> n)))) (a (c <MASK> dist) (a (a (c <MASK> ,) (a (a (c <MASK> *) (v <MASK> m)) (a (a (c <MASK> nadd\_rinv) (v <MASK> x)) (v <PREDICT> n)))) (a (a (c <MASK> *) (v <MASK> n)) (a (a (c <MASK> nadd\_rinv) (v <MASK> x)) (v <MASK> m))))))) (a (a (c <MASK> *) (v <MASK> B)) (a (a (c <MASK> *) (a (a (c <MASK> *) (v <MASK> m)) (v <MASK> n))) (a (a (c <MASK> +) (v <MASK> m)) (v <MASK> n))))))))))))))))))}

Ground truth: \texttt{<START> (num) <END>}
\end{itemize}

\paragraph{Assumptions.}
\begin{itemize}

\item Prompt: \texttt{(<theorem> (a (a (c (fun (bool) (fun (bool) (bool))) ==>) (a (a (c (fun (fun ?1 (bool)) (fun (fun ?1 (bool)) (bool))) =) (a (c (fun (fun ?1 (bool)) (fun ?1 (bool))) GSPEC) (l (v ?1 GEN\%PVAR\%0) (a (c (fun (fun ?1 (bool)) (bool)) ?) (l (v ?1 x) (a (a (a (c (fun ?1 (fun (bool) (fun ?1 (bool)))) SETSPEC) (v ?1 GEN\%PVAR\%0)) (a (a (c (fun (bool) (fun (bool) (bool))) $\wedge$) (a (a (c (fun ?1 (fun (fun ?1 (bool)) (bool))) IN) (v ?1 x)) (v (fun ?1 (bool)) s))) (a (a (c (fun ?0 (fun ?0 (bool))) =) (a (v (fun ?1 ?0) f) (v ?1 x))) (v ?0 a)))) (v ?1 x))))))) (v (fun ?1 (bool)) t))) (a (a (c (fun (bool) (fun (bool) (bool))) ==>) <PREDICT>) (a (c (fun (fun ?1 (bool)) (bool)) !) (l (v ?1 x) (a (a (c (fun (bool) (fun (bool) (bool))) ==>) (a (a (c (fun (bool) (fun (bool) (bool))) $\wedge$) (a (v (fun ?1 (bool)) P) (v ?1 x))) (a (v (fun ?1 (bool)) Q) (v ?1 x)))) (a (c (fun (bool) (bool)) $\sim$) (a (a (c (fun ?0 (fun ?0 (bool))) =) (a (v (fun ?1 ?0) f) (v ?1 x))) (v ?0 a)))))))))}
 
Ground truth: \texttt{<START> (a (a (c (fun (bool) (fun (bool) (bool))) $\wedge$) (a (c (fun (fun ?1 (bool)) (bool)) !) (l (v ?1 x) (a (a (c (fun (bool) (fun (bool) (bool))) ==>) (a (v (fun ?1 (bool)) P) (v ?1 x))) (a (a (c (fun ?1 (fun (fun ?1 (bool)) (bool))) IN) (v ?1 x)) (v (fun ?1 (bool)) s)))))) (a (c (fun (fun ?1 (bool)) (bool)) !) (l (v ?1 x) (a (a (c (fun (bool) (fun (bool) (bool))) ==>) (a (a (c (fun (bool) (fun (bool) (bool))) $\wedge$) (a (v (fun ?1 (bool)) P) (v ?1 x))) (a (v (fun ?1 (bool)) Q) (v ?1 x)))) (a (c (fun (bool) (bool)) $\sim$) (a (a (c (fun ?1 (fun (fun ?1 (bool)) (bool))) IN) (v ?1 x)) (v (fun ?1 (bool)) t))))))) <END>}

Source theorem pretty printed: \texttt{\{x | x IN s $\wedge$ f x = a\} = t  ==> (!x. P x ==> x IN s) $\wedge$ (!x. P x $\wedge$ Q x ==> $\sim$(x IN t))  ==> (!x. P x $\wedge$ Q x ==> $\sim$(f x = a))}

\item Prompt: \texttt{(<theorem> (a (c (fun (fun (fun (cart (real) N) (bool)) (bool)) (bool)) !) (l (v (fun (cart (real) N) (bool)) s) (a (a (c (fun (bool) (fun (bool) (bool))) ==>) <PREDICT>) (a (a (c (fun (fun (cart (real) N) (bool)) (fun (fun (cart (real) N) (bool)) (bool))) =) (a (c (fun (fun (cart (real) N) (bool)) (fun (cart (real) N) (bool))) inside) (v (fun (cart (real) N) (bool)) s))) (c (fun (cart (real) N) (bool)) EMPTY))))))}

Ground truth: \texttt{<START> (a (a (c (fun (bool) (fun (bool) (bool))) $\wedge$) (a (c (fun (fun (cart (real) N) (bool)) (bool)) connected) (a (a (c (fun (fun (cart (real) N) (bool)) (fun (fun (cart (real) N) (bool)) (fun (cart (real) N) (bool)))) DIFF) (c (fun (cart (real) N) (bool)) UNIV)) (v (fun (cart (real) N) (bool)) s)))) (a (c (fun (bool) (bool)) $\sim$) (a (c (fun (fun (cart (real) N) (bool)) (bool)) bounded) (a (a (c (fun (fun (cart (real) N) (bool)) (fun (fun (cart (real) N) (bool)) (fun (cart (real) N) (bool)))) DIFF) (c (fun (cart (real) N) (bool)) UNIV)) (v (fun (cart (real) N) (bool)) s))))) <END>}

Source theorem pretty printed: \texttt{!s. connected ((:real$\hat{}$N) DIFF s) $\wedge$ $\sim$bounded ((:real$\hat{}$N) DIFF s) ==> inside s = \{\}}

\item Prompt: \texttt{(<theorem> (a (a (c (fun (bool) (fun (bool) (bool))) ==>) (a (a (c (fun (bool) (fun (bool) (bool))) $\wedge$) (v (bool) q)) (a (c (fun (bool) (bool)) $\sim$) (v (bool) p)))) (a (a (c (fun (bool) (fun (bool) (bool))) ==>) <PREDICT>) (v (bool) r))))}

Ground truth: \texttt{<START> (a (a (c (fun (bool) (fun (bool) (bool))) =) (v (bool) p)) (v (bool) q)) <END>}

Source theorem pretty printed: \texttt{q $\wedge$ $\sim$p ==> (p <=> q) ==> r}

\item Prompt: \texttt{(<theorem> (a (c (fun (fun (fun (cart (real) N) (real)) (bool)) (bool)) !) (l (v (fun (cart (real) N) (real)) f) (a (c (fun (fun (fun (real) (real)) (bool)) (bool)) !) (l (v (fun (real) (real)) g) (a (c (fun (fun (cart (real) N) (bool)) (bool)) !) (l (v (cart (real) N) x) (a (a (c (fun (bool) (fun (bool) (bool))) ==>) <PREDICT>) (a (a (c (fun (fun (cart (real) N) (real)) (fun (net (cart (real) N)) (bool))) real\_continuous) (a (a (c (fun (fun (real) (real)) (fun (fun (cart (real) N) (real)) (fun (cart (real) N) (real)))) o) (v (fun (real) (real)) g)) (v (fun (cart (real) N) (real)) f))) (a (c (fun (cart (real) N) (net (cart (real) N))) at) (v (cart (real) N) x)))))))))))}

Ground truth: \texttt{<START> (a (a (c (fun (bool) (fun (bool) (bool))) $\wedge$) (a (a (c (fun (fun (cart (real) N) (real)) (fun (net (cart (real) N)) (bool))) real\_continuous) (v (fun (cart (real) N) (real)) f)) (a (c (fun (cart (real) N) (net (cart (real) N))) at) (v (cart (real) N) x)))) (a (a (c (fun (fun (real) (real)) (fun (net (real)) (bool))) real\_continuous) (v (fun (real) (real)) g)) (a (a (c (fun (net (real)) (fun (fun (real) (bool)) (net (real)))) within) (a (c (fun (real) (net (real))) atreal) (a (v (fun (cart (real) N) (real)) f) (v (cart (real) N) x)))) (a (a (c (fun (fun (cart (real) N) (real)) (fun (fun (cart (real) N) (bool)) (fun (real) (bool)))) IMAGE) (v (fun (cart (real) N) (real)) f)) (c (fun (cart (real) N) (bool)) UNIV))))) <END>}

Source theorem pretty printed: \texttt{!f g x. f real\_continuous at x $\wedge$ g real\_continuous atreal (f x) within IMAGE f (:real$\hat{}$N) ==> g o f real\_continuous at x}

\item Prompt: \texttt{(<theorem> (a (c (fun (fun (fun (cart (real) M) (cart (real) N)) (bool)) (bool)) !) (l (v (fun (cart (real) M) (cart (real) N)) f) (a (c (fun (fun (fun (cart (real) M) (cart (real) P)) (bool)) (bool)) !) (l (v (fun (cart (real) M) (cart (real) P)) g) (a (c (fun (fun (fun (cart (real) M) (bool)) (bool)) (bool)) !) (l (v (fun (cart (real) M) (bool)) s) (a (c (fun (fun (num) (bool)) (bool)) !) (l (v (num) n) (a (a (c (fun (bool) (fun (bool) (bool))) ==>) <PREDICT>) (a (a (a (c (fun (num) (fun (fun (cart (real) M) (bool)) (fun (fun (cart (real) M) (cart (real) (finite\_sum N P))) (bool)))) baire) (v (num) n)) (v (fun (cart (real) M) (bool)) s)) (l (v (cart (real) M) x) (a (a (c (fun (cart (real) N) (fun (cart (real) P) (cart (real) (finite\_sum N P)))) pastecart) (a (v (fun (cart (real) M) (cart (real) N)) f) (v (cart (real) M) x))) (a (v (fun (cart (real) M) (cart (real) P)) g) (v (cart (real) M) x)))))))))))))))}

Ground truth: \texttt{<START> (a (a (c (fun (bool) (fun (bool) (bool))) $\wedge$) (a (a (a (c (fun (num) (fun (fun (cart (real) M) (bool)) (fun (fun (cart (real) M) (cart (real) N)) (bool)))) baire) (v (num) n)) (v (fun (cart (real) M) (bool)) s)) (v (fun (cart (real) M) (cart (real) N)) f))) (a (a (a (c (fun (num) (fun (fun (cart (real) M) (bool)) (fun (fun (cart (real) M) (cart (real) P)) (bool)))) baire) (v (num) n)) (v (fun (cart (real) M) (bool)) s)) (v (fun (cart (real) M) (cart (real) P)) g))) <END>}

Source theorem pretty printed: \texttt{!f g s n. baire n s f $\wedge$ baire n s g ==> baire n s (lambda x. pastecart (f x) (g x))}
\end{itemize}

\paragraph{Equalities.}
\begin{itemize}
\item Prompt: \texttt{(<theorem> (a (c (fun (fun (fun ?0 (cart (real) (2))) (bool)) (bool)) !) (l (v (fun ?0 (cart (real) (2))) f) (a (c (fun (fun (fun ?0 (cart (real) (2))) (bool)) (bool)) !) (l (v (fun ?0 (cart (real) (2))) g) (a (c (fun (fun (fun ?0 (bool)) (bool)) (bool)) !) (l (v (fun ?0 (bool)) s) (a (a (c (fun (bool) (fun (bool) (bool))) ==>) (a (c (fun (fun ?0 (bool)) (bool)) FINITE) (v (fun ?0 (bool)) s))) (a (a (c (fun (cart (real) (2)) (fun (cart (real) (2)) (bool))) =) (a (a (c (fun (fun ?0 (bool)) (fun (fun ?0 (cart (real) (2))) (cart (real) (2)))) cproduct) (v (fun ?0 (bool)) s)) (l (v ?0 x) (a (a (c (fun (cart (real) (2)) (fun (cart (real) (2)) (cart (real) (2)))) complex\_mul) (a (v (fun ?0 (cart (real) (2))) f) (v ?0 x))) (a (v (fun ?0 (cart (real) (2))) g) (v ?0 x)))))) <PREDICT>)))))))))}

Ground truth: \texttt{<START> (a (a (c (fun (cart (real) (2)) (fun (cart (real) (2)) (cart (real) (2)))) complex\_mul) (a (a (c (fun (fun ?0 (bool)) (fun (fun ?0 (cart (real) (2))) (cart (real) (2)))) cproduct) (v (fun ?0 (bool)) s)) (v (fun ?0 (cart (real) (2))) f))) (a (a (c (fun (fun ?0 (bool)) (fun (fun ?0 (cart (real) (2))) (cart (real) (2)))) cproduct) (v (fun ?0 (bool)) s)) (v (fun ?0 (cart (real) (2))) g))) <END>}

Source theorem pretty printed: \texttt{!f g s. FINITE s ==> cproduct s ($\backslash$x. f x * g x) = cproduct s f * cproduct s g}

\item Prompt: \texttt{(<theorem> (a (c (fun (fun (fun (cart (real) N) (bool)) (bool)) (bool)) !) (l (v (fun (cart (real) N) (bool)) s) (a (c (fun (fun (fun (cart (real) N) (bool)) (bool)) (bool)) !) (l (v (fun (cart (real) N) (bool)) t) (a (a (c (fun (bool) (fun (bool) (bool))) ==>) (a (a (c (fun (bool) (fun (bool) (bool))) $\wedge$) (a (c (fun (fun (cart (real) N) (bool)) (bool)) convex) (v (fun (cart (real) N) (bool)) s))) (a (a (c (fun (bool) (fun (bool) (bool))) $\wedge$) (a (c (fun (fun (cart (real) N) (bool)) (bool)) affine) (v (fun (cart (real) N) (bool)) t))) (a (c (fun (bool) (bool)) $\sim$) (a (a (c (fun (fun (cart (real) N) (bool)) (fun (fun (cart (real) N) (bool)) (bool))) =) (a (a (c (fun (fun (cart (real) N) (bool)) (fun (fun (cart (real) N) (bool)) (fun (cart (real) N) (bool)))) INTER) (a (c (fun (fun (cart (real) N) (bool)) (fun (cart (real) N) (bool))) relative\_interior) (v (fun (cart (real) N) (bool)) s))) (v (fun (cart (real) N) (bool)) t))) (c (fun (cart (real) N) (bool)) EMPTY)))))) (a (a (c (fun (fun (cart (real) N) (bool)) (fun (fun (cart (real) N) (bool)) (bool))) =) <PREDICT>) (a (a (c (fun (fun (cart (real) N) (bool)) (fun (fun (cart (real) N) (bool)) (fun (cart (real) N) (bool)))) INTER) (a (c (fun (fun (cart (real) N) (bool)) (fun (cart (real) N) (bool))) closure) (v (fun (cart (real) N) (bool)) s))) (v (fun (cart (real) N) (bool)) t)))))))))}

Ground truth: \texttt{<START> (a (c (fun (fun (cart (real) N) (bool)) (fun (cart (real) N) (bool))) closure) (a (a (c (fun (fun (cart (real) N) (bool)) (fun (fun (cart (real) N) (bool)) (fun (cart (real) N) (bool)))) INTER) (v (fun (cart (real) N) (bool)) s)) (v (fun (cart (real) N) (bool)) t))) <END>}

Source theorem pretty printed: \texttt{!s t. convex s $\wedge$ affine t $\wedge$ $\sim$(relative\_interior s INTER t = \{\}) ==> closure (s INTER t) = closure s INTER t}

\item Prompt: \texttt{(<theorem> (a (a (c (fun (bool) (fun (bool) (bool))) ==>) (a (a (c (fun (fun ?0 (bool)) (fun (fun ?0 (bool)) (bool))) SUBSET) (v (fun ?0 (bool)) t)) (a (a (c (fun (fun ?0 (bool)) (fun (fun ?0 (bool)) (fun ?0 (bool)))) DIFF) (c (fun ?0 (bool)) UNIV)) (v (fun ?0 (bool)) s)))) (a (a (c (fun (fun ?0 (bool)) (fun (fun ?0 (bool)) (bool))) =) <PREDICT>) (c (fun ?0 (bool)) EMPTY))))}

Ground truth: \texttt{<START> (a (a (c (fun (fun ?0 (bool)) (fun (fun ?0 (bool)) (fun ?0 (bool)))) INTER) (v (fun ?0 (bool)) s)) (v (fun ?0 (bool)) t)) <END>}

Source theorem pretty printed: \texttt{t SUBSET (:?0) DIFF s ==> s INTER t = \{\}}

\item Prompt: \texttt{(<theorem> (a (c (fun (fun (real) (bool)) (bool)) !) (l (v (real) x) (a (a (c (fun (real) (fun (real) (bool))) =) <PREDICT>) (a (c (fun (real) (real)) real\_abs) (v (real) x))))))}

Ground truth: \texttt{<START> (a (a (c (fun (real) (fun (num) (real))) real\_pow) (a (c (fun (real) (real)) sqrt) (v (real) x))) (a (c (fun (num) (num)) NUMERAL) (a (c (fun (num) (num)) BIT0) (a (c (fun (num) (num)) BIT1) (c (num) \_0))))) <END>}

Source theorem pretty printed: \texttt{!x. sqrt x pow 2 = abs x}

\item Prompt: \texttt{(<theorem> (a (a (c (fun (fun A (bool)) (fun (fun A (bool)) (bool))) =) <PREDICT>) (a (c (fun (fun A (bool)) (fun A (bool))) GSPEC) (l (v A GEN\%PVAR\%0) (a (c (fun (fun A (bool)) (bool)) ?) (l (v A y) (a (a (a (c (fun A (fun (bool) (fun A (bool)))) SETSPEC) (v A GEN\%PVAR\%0)) (a (a (c (fun (bool) (fun (bool) (bool))) $\wedge$) (a (a (c (fun A (fun (fun A (bool)) (bool))) IN) (v A y)) (v (fun A (bool)) s))) (a (a (c (fun A (fun A (bool))) =) (v A y)) (v A x)))) (v A y))))))))}

Ground truth: \texttt{<START> (a (a (c (fun A (fun (fun A (bool)) (fun A (bool)))) INSERT) (v A x)) (v (fun A (bool)) s)) <END>}

Source theorem pretty printed: \texttt{x INSERT s = \{y $\mid$ y IN s $\wedge$ y = x\}}

\end{itemize}

\end{document}